\documentclass{article}
\usepackage[utf8]{inputenc}
\usepackage[margin=1in]{geometry} 
\usepackage{breakcites}
\usepackage{amsmath,amssymb,amsfonts}
\usepackage{acro}
\acsetup{first-style=short,hyperref=true}
\DeclareAcronym{PMx}{
 short = PMx,
 long = Predictive Maintenance
}
\DeclareAcronym{PdM}{
 short = PdM,
 long = Predictive Maintenance
}
\DeclareAcronym{PHM}{
 short = PHM,
 long = Predictive/Prognostic Health Management
}
\DeclareAcronym{CBM}{
 short = CBM,
 long = Condition Based Maintenance
}
\DeclareAcronym{RCM}{
 short = RCM,
 long = Reliability Centered Maintenance
}
\DeclareAcronym{USA}{
 short = USA,
 long = United States of America
}
\DeclareAcronym{EU}{
 short = EU,
 long = European Union
}
\DeclareAcronym{RUL}{
 short = RUL,
 long = Remaining Useful Life
}
\DeclareAcronym{NASA}{
 short = NASA,
 long = National Aeronautics and Space Administration
}
\DeclareAcronym{FEMTO}{
 short = FEMTO,
 long = Franche-Comt\'{e} Electronics Mechanics Thermal Science and Optics
}
\DeclareAcronym{IMS}{
 short = IMS,
 long = Intelligent Maintenance Systems
}
\DeclareAcronym{IGBT}{
 short = IGBT,
 long = Insulated Gate Bipolar Transistor
}
\DeclareAcronym{DET}{
 short = DET,
 long = Digital Engineering Transformation
}
\DeclareAcronym{CNC}{
 short = CNC,
 long = Computer Numerical Control
}
\DeclareAcronym{IRT}{
 short = IRT,
 long = Infrared Thermography
}
\DeclareAcronym{RNN}{
 short = RNN,
 long = Recurrent Neural Network
}
\DeclareAcronym{NGB}{
 short = NGB,
 long = Nose Gearboxe
}
\DeclareAcronym{TPR}{
 short = TPR,
 long = True Positive Rate
}
\DeclareAcronym{FPR}{
 short = FPR,
 long = False Postive Rate
}
\DeclareAcronym{TNR}{
 short = TNR,
 long = True Negative Rate
}
\DeclareAcronym{HUMS}{
 short = HUMS,
 long = Health and Usage Monitoring System
}
\DeclareAcronym{OAT}{
 short = OAT,
 long = Outside Air Temperature
}
\DeclareAcronym{TGT}{
 short = TGT,
 long = Turbine Gas Temperature
}
\DeclareAcronym{NG}{
 short = NG,
 long = Compressor Speed
}
\DeclareAcronym{NP}{
 short = NP,
 long = Power Turbine Speed
}
\DeclareAcronym{IAS}{
 short = IAS,
 long = Indicated Airspeed
}
\DeclareAcronym{LPLQ}{
 short = LPLQ,
 long = Low Power/Low Torque
}
\DeclareAcronym{ELM}{
 short = ELM,
 long = Extreme Learning Machine
}
\DeclareAcronym{GLR}{
 short = GLR,
 long = Generalized Likelihood Ratio
}
\DeclareAcronym{ARL}{
 short = ARL,
 long = Average Run Length
}
\DeclareAcronym{EDD}{
 short = EDD,
 long = Expected Detection Delay
}
\DeclareAcronym{FPCA}{
 short = FPCA,
 long = Functional Principal Component Analysis
}
\DeclareAcronym{RLA}{
 short = RLA,
 long = Randomized Low-rank Approximation
}
\DeclareAcronym{LSTM}{
 short = LSTM,
 long = Long Short-Term Memory
}
\DeclareAcronym{CART}{
 short = CART,
 long = Classification And Regression Trees
}
\DeclareAcronym{COSMO}{
 short = COSMO,
 long = Consensus Self-Organizing Models
}
\DeclareAcronym{RPM}{
 short = RPM,
 long = Rotations Per Minute
}
\DeclareAcronym{VHUMS}{
 short = VHUMS,
 long = Vehicle Health and Usage Monitoring System
}
\DeclareAcronym{ARMA}{
 short = ARMA,
 long = Autoregressive Moving Average
}
\DeclareAcronym{PCA}{
 short = PCA,
 long = Principle Component Analysis
}
\DeclareAcronym{ATM}{
 short = ATM,
 long = Automated Teller Machine
}
\DeclareAcronym{NLP}{
 short = NLP,
 long = Natural Language Processing
}
\DeclareAcronym{UCI}{
 short = UCI,
 long = {University of California, Irvine}
}
\DeclareAcronym{POMDP}{
 short = POMDP,
 long = Partially Observable Markov Decision Process
}
\DeclareAcronym{FMEA}{
 short = FMEA,
 long = Failure Mode and Effects Analysis
}
\DeclareAcronym{FMECA}{
 short = FMECA,
 long = {Failure Modes, Effects and Criticality Analysis}
}
\DeclareAcronym{FVL}{
 short = FVL,
 long = Future Vertical Lift
}
\DeclareAcronym{DSTE}{
 short = DSTE,
 long = Dempster-Shafer Evidence Theory
}
\DeclareAcronym{ICA}{
 short = ICA,
 long = Independent Component Analysis
}
\DeclareAcronym{PSR}{
 short = PSR,
 long = Predictive State Representation
}
\DeclareAcronym{EGT}{
 short = EGT,
 long = Exhausted Gas Temperature
}
\DeclareAcronym{WF}{
 short = WF,
 long = Fuel FLow
}
\DeclareAcronym{N2}{
 short = N2,
 long = Core Speed
}

\usepackage{hyperref}
\usepackage{url}

\usepackage{adjustbox}
\usepackage{array}

\newcolumntype{R}[2]{%
    >{\adjustbox{angle=#1,lap=\width-(#2)}\bgroup}%
    b{1em}%
    <{\egroup}%
}
\newcolumntype{L}[2]{%
    >{\adjustbox{angle=#1,lap=(#2)-\width,raise=1.25cm}\bgroup}%
    b{1em}%
    <{\egroup}%
}

\newcommand{\textcite}[1]{\cite{#1}}%

\title{System-Level Predictive Maintenance:\\ Review of Research Literature and Gap Analysis}
\author{Kyle Miller \and Artur Dubrawski \thanks{Both authors are with the Auton Lab, Carnegie Mellon University. Copyright (c) 2019 Carnegie Mellon University.}}
\date{October 31, 2019}

\begin{document}

\maketitle

\abstract{ 
This paper reviews current literature in the field of predictive maintenance from the system point of view. 
We differentiate the existing capabilities of condition estimation and failure risk forecasting as currently applied to simple components, from the capabilities needed to solve the same tasks for complex assets. 
System-level analysis faces more complex latent degradation states, it has to comprehensively account for active maintenance programs at each component level and consider coupling between different maintenance actions, 
while reflecting increased monetary and safety costs for system failures.
As a result, methods that are effective for forecasting risk and informing maintenance decisions regarding individual components do not readily scale to provide reliable sub-system or system level insights. 
A novel holistic modeling approach is needed to incorporate available structural and physical knowledge and naturally handle the complexities of actively fielded and maintained assets.
}

\tableofcontents

\section{Introduction}
Predictive maintenance describes an approach to equipment management that focuses on exploiting sensing, inspection, and maintenance data to forecast future degradation state, remaining-useful-life, or similar quantity characterizing expected future performance of the equipment. 
Such forecasts are then used to optimize maintenance planning, supply chain, and other maintenance, design, and engineering activities.
As a conceptual framework, it has gained significant popularity in recent years. This is not least due the very attractive claim that predictive maintenance can significantly improve over the state-of-practice by more closely aligning maintenance effort with maintenance need, thereby saving significant amounts of money and time while decreasing unplanned downtime and uncertainty. In applications with limits on equipment availability and/or budgets, predictive maintenance promises to enable intelligent planing to effectively and efficiently satisfy such constraints. 

\begin{figure}
\centering
\includegraphics[width=0.75\textwidth]{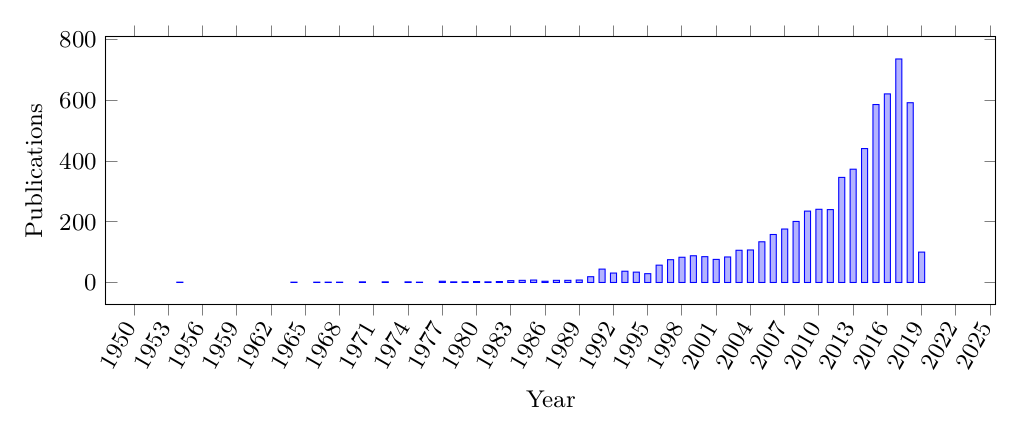}
\caption{Count of predictive maintenance related academic publications by year (as of March 1, 2019).}\label{pubsbyyear}
\end{figure}

Predictive maintenance is sometimes abbreviated \ac{PMx} or \ac{PdM} and sometimes referred to as predictive/prognostic health management (\ac{PHM}). It is closely associated with condition based maintenance (\ac{CBM}) and reliability centered maintenance (\ac{RCM}).
Figure~\ref{pubsbyyear} shows the number of predictive maintenance related academic publications by year\footnote{Web of Science query (conducted on 2019-03-01): \texttt{TS=("predictive maintenance" OR "condition estimation" OR "remaining useful life" OR "degradation model" OR "failure prediction") AND SU=(ENGINEERING OR COMPUTER SCIENCE OR SCIENCE TECHNOLOGY OTHER TOPICS OR MATHEMATICS OR MECHANICS OR ROBOTICS OR OPERATIONS RESEARCH MANAGEMENT SCIENCE )}}. Note the low count in 2019 is an artifact due to the date of the query. Table~\ref{top5countries} lists the top 5 countries of origin by article count. Table~\ref{top10funding} lists the top 10 funding agencies acknowledged by article count. These publication records demonstrate a growing interest in the field, which is likely correlated with advances in machine learning and artificial intelligence, and a reduction in data storage and processing costs over the past decade.

This vast amount of literature makes a comprehensive review challenging. Rather, we review the recent literature on predictive maintenance with a focus on complex equipment at the system and fleet/enterprise levels, examples of which include airlines, truck fleets, etc. The costs, repair time, number of components, variation in use/duty/load, and amount and scope of available data, are all dramatically higher in such scenarios as compared to analysis of individual components. With an increase in the problem complexity and scope, it may not be effective to craft data processing pipelines, data featurizations, and predictive models, for individualized components and failure modes, as is commonly demonstrated in the literature~\cite{rognvaldsson2018self}. 

The remainder of this document is structured as follows. In Section~\ref{sec:complexreview} we provide our view on what differentiates predictive maintenance of complex assets from individual components. In Section~\ref{sec:reviewofreviews} we review other relevant review articles, highlighting recent conceptual trends and industrial foci. In Section~\ref{sec:generalreview} we review primary research, organized by principal concepts in predictive maintenance; condition estimation, forecasting, planning/scheduling, and performance quantification. In Section~\ref{sec:gap} we characterize the gap between prior work and the needed capabilities for system-level PMx and review relevant existing work. In Section~\ref{sec:conclusion} we conclude with promising future directions.

\begin{table}[ht!]
\centering
\caption{Top 5 leading countries by publication count.}\label{top5countries}
{\small
\begin{tabular}{lcc}
Country & \shortstack{Number of\\publications} & \shortstack{\% of\\publications}\\\hline
United States Of America & 1578 & 25.402\\
People's Republic China & 1544 & 24.855\\
France & 421 & 6.777\\
United Kingdom & 327 & 5.264\\
Canada & 245 & 3.944\\
\hline
\end{tabular}
}
\end{table}
\begin{table}[ht!]
\centering
\caption{Top 10 leading funding agencies by publication count, after some de-duplication.}\label{top10funding}
\resizebox{\textwidth}{!}{
{\small
\begin{tabular}{p{0.5\linewidth}ccc}
Agency & Country & \shortstack{Number of\\publications} & \shortstack{\% of\\publications}\\\hline
National Natural Science Foundation of China & China & 561 & 9.034\\
National Science Foundation of China & China & 111 & 1.787\\
National Science Foundation & \ac{USA} & 96 & 1.546\\
Fundamental Research Funds for the Central Universities & China & 93 & 1.498\\
Natural Sciences and Engineering Research Council of Canada & Canada & 39 & 0.628\\
National Basic Research Program of China 973 Program & China & 35 & 0.564\\
China Scholarship Council & China & 34 & 0.548\\
China Postdoctoral Science Foundation & China & 27 & 0.435\\
Fundamental Research Funds for the Central Universities of China & China & 22 & 0.354\\
European Union & \ac{EU} & 18 & 0.290\\
\hline
\end{tabular}}
}
\end{table}

\begin{figure}[ht!]
\centering
\includegraphics[width=0.5\textwidth]{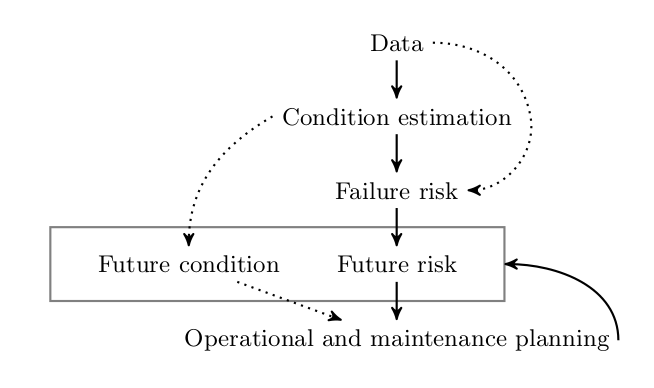}
\caption{Model components for fleet level predictive maintenance of complex equipment}\label{complexmodel}
\end{figure}

\section{Complex Assets}\label{sec:complexreview}
We differentiate between predictive maintenance applied to a single component with that applied to a complex asset.
By complex asset we mean a system of several interacting components. 
In most cases, component interactions such as redundancies make application of predictive maintenance focused on each constituent component, an unsatisfying solution at the system level. Here, we itemize several fundamental distinctions between component level and system level problem elements to emphasize the importance of this differentiation.

\paragraph{Faults.} In single component analysis, faults are typically not enumerated. Rather, failure is the only outcome.  On the other hand, complex assets may present numerous and varied faults, to the degree that novel fault types may be wholly unobserved in training data. Additionally, for complex assets, faults are typically observed at the sub-system or system level. For example, it may be recorded that an engine fails to start, but not that a particular valve gasket has ruptured. Further, it may be that no individual component fails outright, but rather in their degraded state multiple components fail to work together.

\paragraph{Degradation state.} For individual components, degradation state is synonymous with wear and tear, and is tightly connected with remaining useful life (\ac{RUL})\footnote{Perhaps measured in accumulated load and/or use, as opposed to wall clock time.}. It is typical that degradation state is modeled as a single-dimensional quantity that monotonically increases with use. For complex systems this notion must be extended, for example modeling degradation state as a multidimensional vector encoding the wear state of each constituent component. Additionally, the relationship between degradation state and failure can be complex and non-linear. This relationship is the object of study in the discipline of reliability analysis. The use of reliability models such as fault trees and Bayesian networks in predictive maintenance is briefly touched upon in Section~\ref{sec:dependability_models}. 

\paragraph{Data.} Unlike isolated components, it is typically not cost-effective or not feasible to conduct run-till-failure experiments. As a result, observed data are collected from machines operating in a production environment or in the field, which likely includes significant variation in operating loads. These sources of variation may need to be accounted for in predictive models to achieve desired levels of predictive performance.

Degradation state is almost surely not directly observed in complex assets. Direct observation is sometimes assumed for analysis of individual components, but the volumes of data that would be required to directly record degradation state of all of the individual components in a complex asset would likely be prohibitive to say nothing of large sensing array that would be required. Rather, degradation state will be indirectly observed or perhaps partially observed. Additionally, data will typically reflect sub-system behavior rather than individual component state.

\paragraph{Maintenance actions.}
Maintenance can often be ignored in the context of individual components. If training data consist of run-till-failure experiments, maintenance is not performed. In other cases maintenance may consist of replacing a component near failure, in which case the effect of maintenance is to return a part to like-new condition as is often modeled in the literature~\cite{yildirim2016sensorI,yildirim2016sensorII,hao2017controlling,feng2017cooperative}. In training data this can be viewed as censoring observations and maintenance type and effect need not be explicitly considered. In contrast, complex assets may be actively maintained with both repair and replacement of components, impacting degradation state and/or degradation rates, as well as their estimates, in non-trivial ways.   

As a direct result of maintenance, examples of failure may be rare or absent in available training data. In safety-critical systems, components will be serviced or replaced prior to actual failure events. In down-time sensitive applications maintenance may be performed opportunistically during available maintenance windows rather than in correlation with impending failure. 
Therefore, it is important to account for the effects of maintenance during the development of predictive models.

\paragraph{Fleet.}
A fleet of individual components is often treated as a set of identical pieces, the observations of which can be pooled into a single training set. However, with long lived complex assets, individual histories of maintenance, asset-specific usage histories and aging, customization, and modifications, may result in a set of similar but not identical assets. If the degree of similarity is moderate, due to say unit customization, model training procedures will have to be adapted to reflect the resulting subjectivity. Transfer learning or multi-task learning frameworks may be needed for sharing information across the fleet. Additionally, long lived assets may show additional forms of concept drift. For example, replacement parts may be sourced from a new supplier with slightly different tolerances or operational characteristics. In such a situations, historical data may not be perfectly reflective of the current reality.

\paragraph{Planning.} As with fleets of individual components, fleet-level planning for systems requires taking all assets into consideration for making optimal maintenance decisions. This is typically because finite maintenance resources induce a coupling across maintenance decisions for each asset. However, with complex assets, additional couplings always exist between components of a given asset. If maintenance is to be performed on one component, it may induce or block a maintenance window for another component, and may impact the effective duration of jointly performed maintenance actions.

\subsection{Critical Capabilities}\label{sec:critcapabilities}
Generally, predictive maintenance can be described as failure risk forecasting combined with maintenance planning. There are a number of sub-problems that must be solved to realize PMx capabilities. The importance of each of these sub-problems can vary significantly depending on the use case under consideration. The principal components of PMx are described in Figure~\ref{complexmodel}. Data must be collected and curated for use, requiring infrastructure for data collection, storage, and analysis. One common consideration in the literature is how to facilitate data collection through cloud solutions and IoT technology~\cite{meraghni2018post,chukwuekwe2016reliable}, although cloud solutions are not immediately applicable to some asset types or scenarios due to safety and security concerns. Once data is available, it can be used to estimate historical, current, and future condition or failure risk. As noted in Section~\ref{sec:complexreview}, these are not synonymous, though often conflated. Given ones belief of the future risks, operational and maintenance plans can be formulated to optimize global objectives. 

This viewpoint is fundamentally asset-centric. Figure~\ref{complexmodel} does not call out the need to estimate uncertainties in supply chain (e.g.\ shipping lead times, etc.) or in maintenance itself (e.g.\ time to repair).  This focus on asset-centric capabilities is typical of the literature reviewed. The majority of academic research in the PMx field has focused on condition and failure risk forecasting. Maintenance scheduling has been addressed, but to a lesser extent. Operational planning, such as e.g.\ assigning vehicles to delivery routes~\cite{biteus2017planning}, has been briefly touched upon. Cost-benefit analysis of predictive capabilities (e.g.~\cite{busse2018evaluating}) appears also to be currently understudied. 

\section{Review of Reviews}\label{sec:reviewofreviews}

There exist several reviews of \ac{CBM} and \ac{PMx}. Most of these reviews walk the reader through the basic pipeline of data acquisition, processing and feature extraction, modeling and prediction, and finally decision support. 
Usual points under discussion are classes of data types, tools, and techniques that are commonly used. 
We give a brief overview of these reviews here to build out a description of the current state of the field. 

\textcite{si2011remaining} is one of the most frequently cited papers in the field. The authors review several families of \ac{RUL} prediction approaches. The methods are stratified by whether (degradation) state is directly or indirectly observed. For directly observed state, reviewed approaches include regression based models, Wiener processes, Gamma processes, and Markovian models. For indirect observation, the authors describe filtering-based methods, hazard models, and hidden Markov models. 

\textcite{lei2018machinery} gives a recent review of data acquisition and \ac{RUL} prediction. The authors identify four technical processes; data acquisition, health indicator construction, health stage segmentation, and \ac{RUL} prediction. The authors review four commonly used public data sets for \ac{RUL} prediction; The \ac{NASA} turbofan dataset~\cite{saxena2008turbofan}, the \ac{FEMTO} bearing dataset~\cite{nectoux2012pronostia}, the \ac{IMS} bearing dataset~\cite{qiu2006wavelet,lee2007rexnord}, and a milling dataset~\cite{agogino2007milling}. For each dataset~\textcite{lei2018machinery} give a description, list of important properties, and recounts applications. \textcite{eker2012major} additionally describes a Li-ion battery dataset~\cite{saha2007battery}, a Insulated Gate Bipolar Transistor (\ac{IGBT}) dataset~\cite{celaya2009igbt}, and the Vickler dataset~\cite{virkler1979statistical}. Most of these datasets are from the NASA Ames prognostics data repository~\cite{PCoE}, which currently hosts 16 datasets. 
\textcite{lei2018machinery} summarizes performance metrics used in \ac{RUL} prediction and concludes with future challenges including data volume (either limited or overwhelming), handling multiple failure modes, system level \ac{RUL} prediction, and others.

Several reviews focus on Industry 4.0 and the ``digital-twin'' concepts.
The digital-twin is meant to be a ``living model'' which can forecast effectively the behavior (including failure) of its real-world asset counterpart.\footnote{The idea of a digital twin is also found within the U.S.~Depratment of Defense under the banner of the Digital Engineering Transformation (\ac{DET}) and is spelled out in the Digital Engineering strategy by the \cite{digitalEngineering2018}.}
\textcite{liu2018role} describes the development of the digital-twin concept in aerospace, while \textcite{daily2017predictive} review the potential benefits and necessary ingredients of this concept. These authors describe some of the logistical, engineering, and data-science challenges associated with realizing such an ideal. At the heart is the difficulty in fusing multi-physics models, sensing, and ML/data-science technology. Digital twin concepts appear to apply best to clean-sheet engineering efforts, where the digital twins can be built as part of the engineering process. 
For existing complex systems, the engineering effort to reverse engineer digital models can be formidable and cost prohibitive.    

Industry 4.0 represents the thought that modern industrial environments ought to be fully connected and self-aware. 
This includes automatic fault reporting, self-diagnosis, and automatic control of maintenance and production quality, among other capabilities.
\textcite{chukwuekwe2016reliable} gives a very high level description of several interacting trends surrounding Industry 4.0. 
The authors propose (at the conceptual level) a closed loop feedback system for data-driven predictive maintenance. 
They advise that predictive maintenance elements and capabilities be standardized with an emphasis on interoperability. 
The authors suggest predictive maintenance capabilities need to be developed early in the design phase of modern equipment.
\textcite{lee2015cyber,lee2014service} contrast today's technology with Industry 4.0 in which smart sensors and fault detection (today) are replaced with degradation monitoring and \ac{RUL} prediction in self-aware machines. The authors discuss a 5-level architecture that builds from condition monitoring, to prognostics, to fleet level (peer-to-peer monitoring), to decision support, and finally to resilient control systems. 
Aspects of each level are discussed. \textcite{meraghni2018post} proposes a cloud-based architecture for making use of \ac{RUL} and other prognostic results in PMx applications.

Many predictive maintenance reviews are focused on a specific industry, which heavily influences the nature of available data as well as logistic and process constraints in implementation.
Spinning/cutting, \ac{CNC}, and related machinery is a common focal industry~\cite{ferreiro2016industry,yan2017industrial,vogl2019review,sakib2018challenges,trodd1998practical,lee2014prognostics}. \textcite{merizalde2019maintenance} gives a recent review of predictive maintenance in the wind power industry. \textcite{barajas2008real} gives an early review of best practices in predictive maintenance in the automotive industry. The authors describe important enterprise level concepts in the deployment of predictive maintenance and health monitoring strategies. They break down faults along the dimensions of cost/impact and frequency and identify relevant envelopes for reactive, preventative, and predictive maintenance strategies. 
\textcite{enrico2019application} reviews reliability technologies in the aviation industry, and \textcite{gerdes2016effects} surveys potential impact on flight delays. \textcite{blechertas2009cbm} describes a conceptual and procedural map for condition based maintenance approach for rotorcraft. The authors focus on the value of non-destructive analysis techniques, specifically describing the value of vibration, temperature, acoustic emission, electrical signature analysis, and oil/oil debris analysis. 
\textcite{rezvanizaniani2014review} reviews \ac{RUL} prediction for batteries.
Less common industrial focuses include naval~\cite{tambe2015extensible} and hospital operations~\cite{galanreview}.

Other reviews focus on data processing and the IT infrastructure (e.g.\ cloud infrastructure) needed to deploy fleet level or enterprise level predictive maintenance solutions. 
\textcite{wagner2016overview} gives an overview of different data sources that are used in predictive maintenance and the value of each.    
\textcite{schmidt2016context,schmidt2018cloud} review concepts for adding contextual information to condition estimation, using cloud-based technology to facilitate centralized data processing and sharing fleet information. Similarly, \textcite{galar2015context} proposes a hybrid system that considers contextual information for weighing cumulative damage. The objective of the authors' approach is to fuse the capabilities of physics-of-failure modeling with observed data modeling to shore up the weakness of each. For example, observed data and failure/maintenance events can reduce uncertainty in the historic state of a system which physical models can then forecast.
\textcite{patwardhan2016survey} focuses on the technical infrastructure and steps necessary for data preparation in a big data environment.     
\textcite{dragomir2009review} reviews the advantages and disadvantages of physical-model based and data-driven approaches to RUL prediction.   
\textcite{liu2018information} discusses the challenges of transitioning predictive health management systems to complex systems such as the next generation national airspace system. 

Our review differs from the reviews described above in that we aim to highlight the gap between common predictive maintenance strategies and tools, with those needed to achieve the promise of PMx on large scale complex assets. 

\section{Primary Literature}\label{sec:generalreview}
We now turn to a detailed review of literature highlighting each PMx task. Before jumping in however, we pause to discus the issues that arise when different types of data are or are not available. We categorize information as \emph{sensor data}, which describe the current behavior of an asset, \emph{maintenance logs} which describe the actions taken with the intent to extend the utility of an asset or set of assets, and \emph{fault records} which describe observed failures. 

When fault records are unavailable or insufficient in number to support statistical analysis, the PMx effort is called \emph{unsupervised} in contrast to \emph{supervised}. In the supervised setting direct \ac{RUL} prediction is the most common approach. In the unsupervised setting anomaly detection is the principal approach. In contrast, most planning algorithms presume a failure risk forecast.

\subsection{Condition Estimation and Forecasting}

The bulk of the academic literature in the field focuses on condition estimation and forecasting, often specializing to application domain. For this reason, we structure these studies according to domain and primary PMx sub-task; condition estimation and fault detection or \ac{RUL} prediction. We make note of whether techniques are supervised or unsupervised, grouping similar methods together. Finally, we note special cases where sensor data and/or fault records are not available.

\subsubsection{Bearings, spinning, and cutting}
\paragraph{Condition estimation}
\textcite{jia2019rotating} uses a WS-ZHT1 multifunctional rotor test rig to simulate faults, generating supervised data. The authors evaluate infrared thermography (\ac{IRT}) for condition estimation of bearings. They conclude that \ac{IRT} base condition estimation is more effective than traditional vibration based methods. Other authors working in the supervised setting focus on vibration data.
\textcite{sezer2018industry} fit a model to predict roughness from vibration and temperature data in \ac{CNC} machines. 
\textcite{kateris2014machine} performs condition monitoring of bearings in rotating machinery using vibration data. The authors use neural networks to identify and locate (inner/outer race) faults using fully labeled data collected on a test machine. 

\textcite{ferreiro2016industry} describes unsupervised predictive maintenance in the spinning tool setting. The authors use a finger-print learning method using supervised data to train a fault detector and an envelope analysis to detect outliers.  
\paragraph{\ac{RUL} prediction}
Supervised \ac{RUL} prediction methods for bearings or rotating machinery very often use data from run-till-failure bench-top experiments. Occasionally, partially damaged bearings will be used to accelerate failure in order to gather more failure examples or explore specific failure modes.
\textcite{yan2017industrial} describes a data processing pipeline for predictive maintenance in the industrial setting. The authors demonstrate predicting tool wear and tool \ac{RUL} on \ac{CNC} machines. While the authors describe processes for creating structured data from semi-structured or un-structured data common in the industrial setting, their demonstration focuses on the use of featurized vibrations data via envelope analysis and similar strategies. 
\textcite{li2019remaining} developed a state-space based \ac{RUL} prediction algorithm that is robust to varying operating conditions. The approach uses a particle filter with linear drift term. The linear term is modulated by operating conditions. A pair of operating condition dependent jump coefficients are introduced to the observation model, to account for jump discontinuities or change points in the observed degradation signal. In earlier work, 
\textcite{bian2015degradation} model degradation in a randomly-evolving environment modeled as a continuous-time Markov chain. The authors argue that most hazard models and prior research considers only static environments, which can lead to model-mismatch and degraded model performance if environments do vary. 
\textcite{hao2017residual} consider a serial processing line in which tool wear impacts production quality and production quality of preceding steps effects tool wear rates of downstream steps. 
A linear relationship between burr size and tool wear is presumed. The approach is demonstrated on simulated data.  
\textcite{fumeo2015condition} uses an online-support vector regression machine to efficiently learn/predict \ac{RUL} on railway bearings. The authors use vibration and temperature as inputs from run-to-failure data. 
\textcite{liao2016enhanced} develop feature extraction capabilities for improved \ac{RUL} prediction on bearing systems, again using run-till-failure experiments.  
\textcite{luo2019early} demonstrates \ac{RUL} and wear prediction on spinning tools. The authors use advanced dynamic identification techniques to process vibrations data coupled with deep learning methods to produce their final predictive model.  
\textcite{fang2018real,fang2019image} develops tensor-based methods for \ac{RUL} prediction from streams of infrared images of bearings. When stacked, these images form a rank-3 tensor. The approach is to project the tensors to a low-dimensional tensor subspace and then apply a penalized location-scale regression using \ac{RUL} as the dependent variable. 
\cite{guo2017recurrent} apply recurrent neural networks to \ac{RUL} estimation on bearings in the supervised setting. The authors conclude that \ac{RNN}s give superior performance to self organizing maps.

\textcite{kanawaday2017machine} analyzed industrial cutting tools. The authors used unsupervised techniques to establish outliers. They then trained supervised models to forecast the occurrence of these outliers.  

\subsubsection{Gearboxes}
\paragraph{Condition estimation}
\textcite{zhao2019multiple} develops supervised methods for fusing wavelets and deep learning approaches. The authors demonstrated their method on planetary gearbox fault diagnosis.
\textcite{wade2017applying} use vibration data to estimate condition of nose gearboxes (\ac{NGB}s). Authors cite prior work indicating that vibration exceedences are a sub-optimal condition estimator due to variation in individual aircraft and components. The authors develop aerospace specific metrics for model selection. Data represents 600 assets with 40 ground-truth faults. Authors describe several metrics including bookmakers informedness (\ac{TPR}-\ac{FPR}), historical based \ac{TNR}, asset based \ac{TNR}, in-sample informedness, cross-validation informedness, absolute difference between in-sample and cross-validation informedness, and position shuffle.   
\textcite{wade2015machine} describes data preparation for health status prediction of engine output gearboxes and turbo shaft engines. Health and Usage Monitoring System (\ac{HUMS}) data are used as predictors, including Outside Air Temperature (\ac{OAT}), Turbine Gas Temperature (\ac{TGT}), Torque, Compressor Speed (\ac{NG}), Power Turbine Speed (\ac{NP}), Anti-Ice, Indicated Airspeed (\ac{IAS}), and Barometric altitude. The predictive target is engine removal events for reason of low power/low torque (\ac{LPLQ}). 

\textcite{oehling2019using} suggest that the state-of-art for informing safety from flight data is to monitor for exceedences of established thresholds. The authors use unsupervised outlier detection to identify potentially safety-relevant occurrences from flight data and compare to the exceednce-based approach. Outlier detection is shown to have good utility.   
\paragraph{\ac{RUL} prediction}
\textcite{martin2019dictionary} demonstrates an unsupervised dictionary learning based approach for faults in wind farms. Data are gearbox vibration records for six turbines (publicly available). Condition evaluation is effected by building anomaly detection capability using learned dictionaries by means of a ``dictionary distance.'' Dictionaries are realized as a sparse coding model.   
 
\subsubsection{Turbines}
\paragraph{Condition estimation}
\textcite{rahman2018diagnostics} use a supervised signature based algorithm for detecting and characterizing faults. Fault signatures are produced by simulating different fault types. \cite{rausch_integrated_2007} also used supervised learning to detect and classify faults and used these classifications to adjust flight parameters in real-time for improved flight safety. Training data was again produced using numeric simulations.

\textcite{yan2016one} uses unsupervised anomaly detection of redundant (simultaneous) temperature measures to diagnose combustor issues in gas turbine engines. Data is sampled at 1/60 Hz and an extreme learning machine (\ac{ELM}) is adapted for anomaly detection. \textcite{michelassi2018machine} presents a very similar work. 
\textcite{michau2018fleet} uses deep-learning based anomaly detection methods to identify potential faults in gas turbine data. The authors also explore the use of ``sub-fleets'' creating appropriate cohorts for comparison.    
\paragraph{\ac{RUL} prediction}
\textcite{xue2008instance} developed a fuzzy-similarity based method for estimating RUL on aircraft turbine engines. Authors analyzed cases of high pressure turbine shroud burn faults. Their algorithm identifies peer groups based on exhausted gas temperature (\ac{EGT}), fuel flow (\ac{WF}), and core speed (\ac{N2}) after correcting for flight envelopes. Observed \ac{RUL} from from identified peers is then aggregated to estimate the \ac{RUL} of a target engine.  
Many supervised \ac{RUL} prediction studies use the \ac{NASA} turbofan dataset~\cite{saxena2008turbofan} as benchmark. 
\textcite{fang2017multistream} develop methods for improved multivariate \ac{RUL} regression, including feature selection. 
\textcite{cao2018multi} proposes a change point detection modeling a (linear) gradual degradation to a subset of sensor streams ($p_0$), where observations before and after the change point $k$ are assumed to be i.i.d. normal. The detection is based on a generalized likelihood ratio (\ac{GLR}) statistic considering average run length (\ac{ARL}) and expected detection delay (\ac{EDD}). Several extensions of the technique are proposed e.g. non i.i.d. case and modelling adaptive subset of crushed sensors $p_0$. The method is demonstrated on stock bidding trend detection as well as the \ac{NASA} turbofan dataset.
\textcite{fang2017scalable,fang2018predictive} uses functional Principal Component Analysis (\ac{FPCA}) and location-scale regression are used to predict time to failure (\ac{RUL}) of partially degraded equipment. A multivariate \ac{FPCA} and hierarchical \ac{FPCA} is used for data fusion on a massive dataset.  One of the key contributions is that the scalability of (multivariate) \ac{FPCA} is enhanced by exploiting Randomized Low-rank Approximation (\ac{RLA}) without knowing the rank of the \ac{RLA} in advance. 
\textcite{zhang2018deep} use a 3-layer \ac{LSTM} for gas turbine engine \ac{RUL} prediction. The authors define a health index as the output of a single ReLU neuron, fit to predict 1 at the beginning of an engine's time series and 0 at time of failure, regularized against first differences. This regularization encourages smooth health index trajectories. Finally, the 3-layer \ac{LSTM} is trained for a one-step forecast task. By repeated forecasts at test time, \ac{RUL} is inferred.
\textcite{ragab2016remaining} uses a discrete logic approach for \ac{RUL} prediction given observed operating parameters and condition indicators.    
\textcite{li2019degradation} describes an ensemble \ac{RUL} prediction approach, using Random Forest, \ac{CART}, \ac{RNN}s, and several other algorithms as constituents of the ensemble. The authors show that the ensemble is able to predict \ac{RUL} on the \ac{NASA} turbofan dataset with high accuracy.

\textcite{coraddu2016machine} describe a simulation experiment for \ac{CBM} on gas turbines in naval ship propulsion. A sophisticated physics model characterizing gas turbines and ship propulsion is used to simulate data. They then use ML models to estimate decay rates from observed data. \cite{baraldi_kalman_2012} use ensemble methods to \ac{RUL} of the turbine blades of generators within nuclear power plants using simulated mechanical stress (mechanical fatigue) of the turbine blades.

\subsubsection{Vehicles}
\paragraph{Condition estimation}
\textcite{atamuradov2018railway} describes supervised health indicator (HI) construction, assessment, and prognostics for railway applications.

\textcite{rognvaldsson2018self} describes a life-long learning approach to fault detection, arguing that it is economically infeasible to use human experts to build, evaluate, and field predictive models for each failure mode. This is especially true for novel failure modes or (potentially) occasionally modified equipment. The paper gives a good review of unsupervised methods. The authors remark that most of the prior work presumes high-quality features are provided (presumably by experts) and that little work in the unsupervised space accounts for inter-asset variation. The authors' approach is based on Consensus Self-organizing Models (\ac{COSMO}), and the basic elements of the strategy are to first identify interesting (in an information theoretic sense) functional transformation of raw sensor data and then to compare these derived values across the fleet. One or a few outliers were there is general consensus otherwise in one or more derived signal suggests a fault. The authors conclude that there is a significant need to improve the quality of date in service records.  
\textcite{dubrawski2011techniques} demonstrate detection of escalating maintenance issues by comparing event counts with historical counts as well as with similar cohorts.   
\paragraph{\ac{RUL} prediction}
\textcite{bonissone2005predicting,bonissone2005fuzzy} present a fuzzy-similarity based approach to identify peer groups for a fleet of 1100 locomotives. Peers are similar in maintenance history, usage, and expected behavior. Observed \ac{RUL} is aggregated across peers to estimate \ac{RUL} for target locomotives. The authors use an evolutionary framework for model optimization to maintain an up-to-date similarity measure. 
\textcite{teixeira2015probabilistic} models the evolution of fault magnitude in components, using their supervised model to disregard apparent faults that do not follow expected evolutionary behavior. The result is that their model successfully disregards most cases of ``no fault found.''
\textcite{le2017condition} use ML models to predict \ac{RUL} for engine oil in land based military vehicles. Data collected from \ac{VHUMS} included engine \ac{RPM}, temperature, throttle position, oil temperature, among others. Oil condition was measured by means of laboratory tests. Data included 16 vehicles with a total of 30 oil test results. Rule-learning gave very good performance in stratified cross-validation (number of folds was unspecified).    
\textcite{nascimento2019fleet} proposes an \ac{LSTM} with monotonic damage accumulation. The utility of the model is demonstrated by synthetic simulation. Training data are ``far field stresses,'' and labels are periodic inspections for cracks.
\textcite{magargle2017simulation} gives and in-silico demonstration of digital-twin methodology in support of predictive maintenance for automotive breaks. By reference to the digital twin, wear rate is inferred from data and \ac{RUL} predictions are made.    
\textcite{prytz2015predicting} describes the application of predictive maintenance to a fleet of trucks. Three years of data are used to demonstrate the approach. The authors describe common difficulties; data is co-opted for mining, maintenance records are incomplete and free-text based, etc. The authors note that the feature distribution used for predicting future faults is age dependent, and apply modeling strategies to correct for equipment age. They also discuss issues arising of dependence among observations in the data set and recommend a leave-one-vehicle out cross-validation approach.   
\textcite{nixon2018machine} describes predictive maintenance analysis on diesel engines for military vehicles. Input data consists coarsely sampled measures from the engine management computer. Predictive targets were created by grouping unscheduled maintenance actions by failure mode.     
\textcite{baptista2019remaining} studies how Kalman filtering can be used to smooth \ac{RUL} estimates over time, reducing noise and improving overall accuracy. 

\textcite{cipollini2018condition} evaluate several ML approaches for engine health analysis on naval vessels. The authors conclude that unsupervised anomaly/outlier detection methods are the most appropriate as they can be realized with minimal ground-truth. A public dataset is available for this work.  

\subsubsection{Industrial plant operations}
\paragraph{Condition estimation}
\textcite{amruthnath2018research} explores some unsupervised methods for fault detection using vibration data from a cooling fan.  
\textcite{grass2019unsupervised} proposes an unsupervised approach for anomaly detection in time-series data representing configuration-based electronics production lines.   
\textcite{hendrickx2018fleet} describes an unsupervised clustering approach for comparing similar machines in industrial environments. Anomalies are detected by means of monitoring similarity among machine equivalence classes.     
\textcite{kroll2014system} describes an anomaly detection strategy based on discrete-continuous hybrid automata.  
\paragraph{\ac{RUL} prediction}
\textcite{mattes2012virtual} evaluates Bayesian networks, Random Forest, and linear regression for supervised \ac{RUL} prediction using equipment specific sensor data.    
\textcite{susto2015machine} presents a supervised model for predicting failure within $m$ iterations using data from run-to-failure experiments for an ion implanter tool. 
\textcite{susto2016dealing} explores the application of a time-series featurization approach to \ac{RUL} prediction.
\textcite{bastos2014application} describes an ML framework for predictive maintenance of a nuclear plant. Features consist of monitoring data reported at 1 Hz frequency. The prediction target are failures, recorded in maintenance records. 

\subsubsection{Other}
\paragraph{Condition estimation}
\textcite{poosapati2019enabling} proposes a rule-based strategy for processing anomalies in predictive maintenance applications. Their goal is to develop cognitive reasoning capabilities that can recognize patterns and suggest courses of action.    
\paragraph{\ac{RUL} prediction}
\textcite{cristaldi2016comparative} evaluates a few models for supervised \ac{RUL} prediction of a ``fleet'' of circuit breakers. Models have access to an observed health condition (HC), and forecast the point at which the HC reaches the end-of-life level using observed time-series as inputs. Fleet level data is used to learn probability distributions over HC variation.   
\textcite{cline2017predictive} review 19 years of inspection data for swivels and valves on oil and gas equipment. Authors noted that they were unable to compute residual life of the majority of components due to the fact that most components never failed. Failure within next year was selected as the most viable target. Features included wear-index, derived values thereof, and counts of previous failure or inspection events. 
\textcite{bey2009practical} suggests that data engineering and feature selection through case studies can be used to facilitate later development of prognostic models. The authors demonstrate \ac{RUL} prediction on copy machines.  
\cite{mishra2018bayesian} apply hierarchical Bayesian modeling to forecasting battery performance. The hierarchical modeling structure effects a peer-to-peer comparison, and can make predictions without sensing data based on a battery's peer group (i.e.\ its prior).
\textcite{xin2017dynamic} extends Bayesian hazard modeling for fire and industrial accidents to include dynamics.   


\subsubsection{No sensors}
In the absence of sensor data, researchers have used maintenance and/or failure data to uncover patterns that can forecast future failures.

\textcite{baptista2018forecasting} proposes an \ac{ARMA} based model for supervised prediction of \ac{RUL}/failure risk aimed at reducing unnecessary removals and avoiding failure.  The authors use an \ac{ARMA} model and \ac{PCA} to featurize time-series of past removal/failure events and pass this through a predictive model which forecasts \ac{RUL}. They demonstrate on a data set of 584 engine bleed valve removals.
\textcite{sipos2014log} uses distribution-classification to predict upcoming failure from the distribution of observed fault codes in log-data collected from medical equipment. Service notifications are used to denote failure.    
\textcite{korvesis2018predictive} parse post-flight event logs to predict landing gear faults in aircraft. 
\textcite{kraisangka2016making,kraisangka2018bayesian} show how Bayesian networks can be used to model hazard rates, leading to more powerful models. 
\textcite{wang2017predictive} demonstrate a classification based approach for predictive maintenance in automated teller machines (\ac{ATM}s). The authors use statistics of error message occurrences, occurrences of temporal patterns of error messages, and individual machine characteristics (model, installation date, etc.). Error message type is extracted from error codes present in the \ac{ATM} log files.  Labels are determined by the occurrence of a maintenance ticket.

\textcite{salo2018value} present a poster describing an \ac{NLP} pipeline for extracting useful information from free-text maintenance write-ups in wind farm data. The approach cluster text descriptions into equivalence classes, grouping write-ups that describe the same/similar maintenance actions.     

\subsubsection{No sensors and no faults} If neither explicit failures nor sensor data are available, one can predict future maintenance using historical maintenance.
For example, \textcite{gardner2017driving} uses tensor decomposition to data-mine maintenance data for patterns. A rank-3 tensor is created out of vehicle ID, maintenance action type, and time. An \ac{LSTM} is trained to forecast maintenance actions as well, treating each vehicle's time series as an observation.

   
\subsection{Maintenance and Operational Planning}

Most work in predictive maintenance does not consider variable workloads, operating conditions, or equipment use. Further, those that do explicitly take these issues into account~\cite{hao2017controlling,li2019remaining,bian2015degradation} generally do not forecast use and/or modify usage plans in consideration of degradation status. 
\textcite{biteus2017planning} is an exception. The authors describe an end-to-end predictive maintenance program that predicts failure risks, schedules maintenance actions, and creates condition-aware plans of utilization (route planning) for a fleet of trucks. Maintenance actions are broken down into the smallest possible units and transformed into constraint rules. A random forest is used to predict failure risk, and constrained optimization strategies are used to produce maintenance and route plans. The approach is demonstrated on a fleet of 80,000 trucks and a single component (air dryer purge valve) for which failures are observed in 1.6\% of records. Data are publicly available in \ac{UCI} repository~\cite{Dua:2019}.    

\subsection{Maintenance scheduling}
\textcite{maillart2006maintenance} applies a \ac{POMDP} framework, assuming that without maintenance, system state degrades stochastically, over discrete states, according a known transition function. Maintenance costs are differentiated according to whether they are preventive or reactive. Both types of actions are assumed to return the system to like-new condition.
\ac{POMDP} formulations are also explored by \cite{ghasemi2007optimal,jiang2015pomdp,liPozzi2019predicting}.

\textcite{yildirim2016sensorI,yildirim2016sensorII} represent a two-part paper. In part I, the authors assume the ability to observe a degradation signal which is given by a parametric degradation function plus additive noise.
Observation of the degradation signal allows inference of the asset-specific degradation parameters, some of which are shared across a fleet. A Bayesian model is presumed, and the distribution of \ac{RUL} for each asset is inferred from the observed degradation signal. A cost function relates \ac{RUL} to cost by dictating a different (lower) cost for planned maintenance than for failure events. A maintenance action (planned or otherwise) is presumed to return the asset to ``new'' status (note assets are treated as single-component systems). A mixed-integer program is defined for characterizing total maintenance costs. A constraint on labor capacity couples maintenance actions across assets. In part II, the mixed-integer program is extended to include constraints on asset commitments and loads. These can encode constraints on the number of up/down transitions, total availability or capacity, etc. Experiments demonstrate significant improvement in reliability and reduced costs over standard practice.     
\textcite{yildirim2017integrated} demonstrates a very similar approach to \cite{yildirim2016sensorI,yildirim2016sensorII} for a fleet of wind turbines. The authors add constraints that limit location visits on the part of the maintenance crew, constraints of maintenance effort, and constraints on turbine output which couples the maintenance effort across turbines encouraging concurrent maintenance actions. This leads to cost optimization.    

\textcite{basciftci2018stochastic} extends the mixed-integer programming planning algorithm of \cite{yildirim2016sensorI,yildirim2016sensorII} to include a probabilistic constraint on availability. This constraint ensures that the likelihood of too many assets in maintenance simultaneously is low. The purpose of this constraint is to guard against the risks and costs of unexpected failures.    
\textcite{moghaddass2018joint} solves the joint condition estimation and maintenance planning problem for single-component systems. The authors assume preventative maintenance is less expensive than failure and maintenance actions require a certain lead time. The approach is demonstrated with numerical simulations.    
\textcite{yang2008maintenance} describes a genetic-algorithm optimization approach for scheduling maintenance actions based on noisy \ac{RUL} predictions.    
\textcite{rajora2018intelligent} is a dissertation largely focusing on solving hierarchical coupled constraint optimization problems that arise in maintenance scheduling and assembly planning problems.  
\textcite{hao2017controlling} presumes that system degradation is a function of workload (increased workload increases degradation). The authors develop a control system that dynamically modulates workload between multiple machines, based on posterior degradation belief distributions. The controller seeks to guide failure of machines in such a way that they do not overlap, reducing risk of work-stoppage. The approach is demonstrated on simulated stamping machines.    

\textcite{lin2018multi} argues that most \ac{CBM}-oriented research in the aerospace domain focus on minimizing cost or maximizing availability of single aircraft in isolation, and rarely consider both objectives simultaneously much less that for an entire fleet. The authors propose a model for doing just that. The model assumes a simple deterministic damage function (of time) and cost function. The authors use support vector regression to effect the multi-objective optimization.  
\textcite{feng2017cooperative} describes a learning game-theoretic approach to fleet-level maintenance strategy aimed a minimizing cost under an availability constraint. The game is focused on learning strategies of when to replace line-replaceable modules, given failure probabilities. The authors also touch on the NP-hard nature of the fleet level \ac{CBM} problem. 
\textcite{feng2017heuristic} extends this work to include dispatched and standby sets of aircraft. Again, game theory is used to search for optimal decision strategies.

\subsection{Performance quantification}
It is advisable to understand the level of predictive performance necessary for a predictive maintenance effort to yield positive utility. Such measures serve the important function of defining success both for proofs-of-concept predictive models and system performance while scaling solutions to the enterprise level. 
Toward that end, \textcite{busse2018evaluating} demonstrates an a priori cost-benefit-analysis for predictive maintenance capabilities. This is significant, as such analyses can provide the aforementioned understanding. The authors use a Wiener process with linear drift to model the predictions of a hypothetical \ac{RUL} prediction module. They then push sampled predictions through different maintenance planning strategies, and compute total costs using a hypothetical cost model. The demonstration is conducted for a single component machine with single failure mode. 

\textcite{lei2018machinery} reviews performance metrics for \ac{RUL} prediction. The authors divide metrics into offline and online measures. Offline metrics measure accuracy of \ac{RUL} estimations or failure risks for example. THe proposed online metrics, in contrast, do not require knowledge of future failures, comparing the current \ac{RUL} estimate to its recent estimates.

\subsection{Supply planning}
\textcite{boev2019adaptive} sketch out a constrained optimization based approach for prescribing maintenance plans and spare part availability.

\section{Gap Analysis}\label{sec:gap}
In the reviewed literature, the asset under study is sometimes simple such as a bearing or cutting tool and sometimes complex such as a gas turbine or automotive engine. However, when complex assets are considered, it is largely the case that either only a small number of simple components or a small number of failure modes are studied. As such, these complex assets are treated using methods analogous to those for individual components. This approach has the advantage that methods and insights developed using run-till-failure bench experiments on bearings say, can be utilized on larger systems where run-till-failure is not realistic. 
Further, it could be argued that one could repeat such a process for all the major components and/or failure modes of a complex asset. 
It has been pointed out however, that this approach could be prohibitively expensive due to the resources needed to build and maintain the numerous required models~\cite{rognvaldsson2018self}. Further, if dependencies between failure modes are to be taken into account, then there is little justification for not starting with a comprehensive model approach.

In our view, the primary gaps between our view of PMx for complex assets and current literature, center on the handling of condition and failure risk estimation/forecasting. We identify two principal gaps: failure to incorporate inter-component interactions, and failure to address the effects of maintenance.

\subsection{Modeling Interactions}
Modeling interactions between components can enable sub-system or system level models of failure risk, facilitating PMx for fleets of complex assets. This is not a new concept. We review some initial work in detail below. But first, we highlight work from Dependability Modeling and Analysis, a closely related discipline that specializes in this area.  

\subsubsection{Dependability Modeling and Analysis}\label{sec:dependability_models}
Reliability and dependability analysis is standard practice is product design. 
The term describes the problem of quantifying the risk and nature of failures of (typically complex) equipment. Generally, the goal of dependability modelling is to relate basic events, which often represent failure of individual components, to overall sub-system and system level behavior. Such models can be used to determine the criticality of different components, overall system robustness, as well as to diagnose, correct, and avoid failures. Common methods include Failure Mode and Effects Analysis (\ac{FMEA}), Failure Modes, Effects and Criticality Analysis (\ac{FMECA}), (dynamic) fault-trees, (dynamic) Bayesian networks, and stochastic Petri-nets. These methods are currently being integrated into the DoD digital engineering strategy \cite{boydston_joint_2015} on the Future Vertical Lift (\ac{FVL}) program. Of particular note on the \ac{FVL} efforts is the use of modeling at both the system and subsystem level.

\textcite{chemweno2018risk} gives a recent review of dependability modelling with a focus on the treatment of uncertainty, both uncertainty of predictions (aleatory) and uncertainty of the model (epistemic)~\cite{fox2011distinguishing}.  The authors find that dynamic fault-tree analysis and dynamic Bayesian networks are the most common methods, together accounting for 44\% of the dependability modelling literature (as measured by count of articles). They note that while Bayesian methods are naturally suited for combining evidence from different sources, limited reliability data necessitates quantifying the epistemic uncertainty beyond typically analysis of posterior distributions. Toward that end, the authors review fuzzy analysis, interval analysis, and Dempster-Shafer evidence theory (\ac{DSTE}) for quantifying epistemic uncertainty. \ac{DSTE} is the most common such method accounting for 46\% of articles that address epistemic uncertainty. Finally, the authors identify inclusions of predictive models of failure probability into reliability models as a key future research direction.

In that respect, there are several degrees of potential integration between these distinct modeling exercises. One may build \ac{RUL} and/or failure risk forecasting capability for individual components, treating each as independent. The forecast risks can then be fed into a reliability model to more comprehensively inform risk assessment process. \textcite{lee2019evaluating} can be viewed as a step in this direction. The authors combine estimates of failure probability via a Markov model with a reliability model using a tree-structured Bayesian network. 
If significant dependencies exist in the failure risks of basic events, they will have to be taken into account. Sub-system or system level faults may impact the degradation rates of components (e.g. adjusting workloads or operating conditions due to a fault). In such circumstances is may be desirable to model basic failure risk and system reliability jointly. This could be accomplished using dynamic Bayesian networks, for example. \textcite{chiacchio2016shyfta,chiacchio2016stochastic} join stochastic hybrid automaton with dynamic fault-trees to jointly model age of components and failure risk under dynamic operating conditions. However, no learning is performed as the governing equations of the approach are given upfront. 



\subsubsection{Primary literature focused on complex assets}
Some authors have begun to address the challenges that arise when considering complex assets. Often this means modeling the relationship between sensing and component state and component-component interactions.

\textcite{rodrigues2017remaining} introduces a particle filter model wherein the observation function is informed by system architecture. Incorporating this system-level model allows the method to relate system level performance indicators to component health state. The authors model the component health state as a gamma process. The method is demonstrated on two simulated data sets; a simplified multi-pump hydraulic system and  a multi-component air conditioning system.

\textcite{lee2019evaluating} assume the degradation state of each component is described by a discrete vector with $h_i\in\{0,1,\dots,f_i\}$, where $i=1,\dots,N$ enumerates components and $f_i\in\mathbb{N}$ is the failure state for component $i$. These so-called health states are presumed to be increasing in severity, until failure. Health state values are forecast $n$ time steps into the future using a Markov model, for which the transition matrices $P_{i}$ are known (or learned from historical data) for each component. The $P_{i}$ also encode the assumption of non-decreasing state transitions, i.e.\ no spontaneous repair. Let $\vec{h}_i$ be a one-hot vector encoding the current health state for component $i$, then $P_{i}^n \vec{h}_i$ is the posterior health state distribution for component $i$. Finally, probability of the system or a sub-system level failure is computed by a tree-structured Bayesian network, for which the parameters (conditional probability tables) are presumed known a priori.

\textcite{barde2019optimal} demonstrates a classical reinforcement learning strategy for maintenance of a fleet of trucks. The authors consider 8 components, and use a model-free approach with tabular $Q$ function to learn the optimal maintenance policy under different choices of reward function. This type of reinforcement learning does have optimality guarantees in the limit of sufficient state-action space exploration. However, the main advantage may be that it is easy to integrate complex logistics and incorporate the effects of multiple concurrent maintenance actions. Unfortunately, this kind of approach can only work if (i) the state-action space is discrete and of low enough arity that it can be sufficiently explored, (ii) ample observed data or realistic simulations of equipment histories are available, and (iii) failures are observed. If maintenance is largely preventative, the learning agent will not effectively be able to directly learn policy since it will not encounter penalties associated with failure. Additionally, in real-world complex equipment, the state-action space is likely to be at least partly continuous and complex, requiring function approximation techniques to learn the $Q$ function. In practice, these conditions would require massive amounts of trials to find good policies. Further, current opinion in the field is that reinforcement learning using model approximation can be very difficult to tune properly and can produce sporadic unanticipated behavior. This is unacceptable in safety critical applications such as e.g.\ aerospace.

\textcite{lin2018multi} focuses on maintenance planning for a fleet of aircraft. The authors presume that a probability-of-failure model is given for each component, which is a function of the component's damage level. Aircraft failure probability is taken as the maximum component level failure probability. This assumption may be in error and the aircraft failure probability depends on the statistical dependency between components. In any case, the authors define a repair cost function, dependent on the damage level of a component and a wasted \ac{RUL} function. Finally, they optimize a two-objective decision model under the constraint that failure probability is very small.

\textcite{hao2015simultaneous} consider sub-system level sensing, e.g.\ vibration measurements, and study how one can isolate component-level degradation signals. The authors use independent component analysis (\ac{ICA}) to separate the degradation signals for a known number of components and demonstrate \ac{RUL} prediction on synthetic data.
\textcite{blancke2018predictive} describes the use of Petri-nets for failure risk forecasting on complex systems. Their approach relies on expert knowledge of failure physics, and models fault propagation using a colored Petri net. Modeling the fault propagation allows for prescriptive diagnostic inference as well.

\subsection{Maintenance}
Maintenance of complex assets raises two primary issues. The first is that maintenance censors future failure events. Second, maintenance actions could alter the latent degradation state and its trajectory in non-trivial ways.
Yet, little to no work has been put toward modeling the impact of maintenance on the latent degradation state.

\begin{figure}[ht!]
\centering
\includegraphics[width=0.33\textwidth,trim=0mm 0 0 0mm, clip]{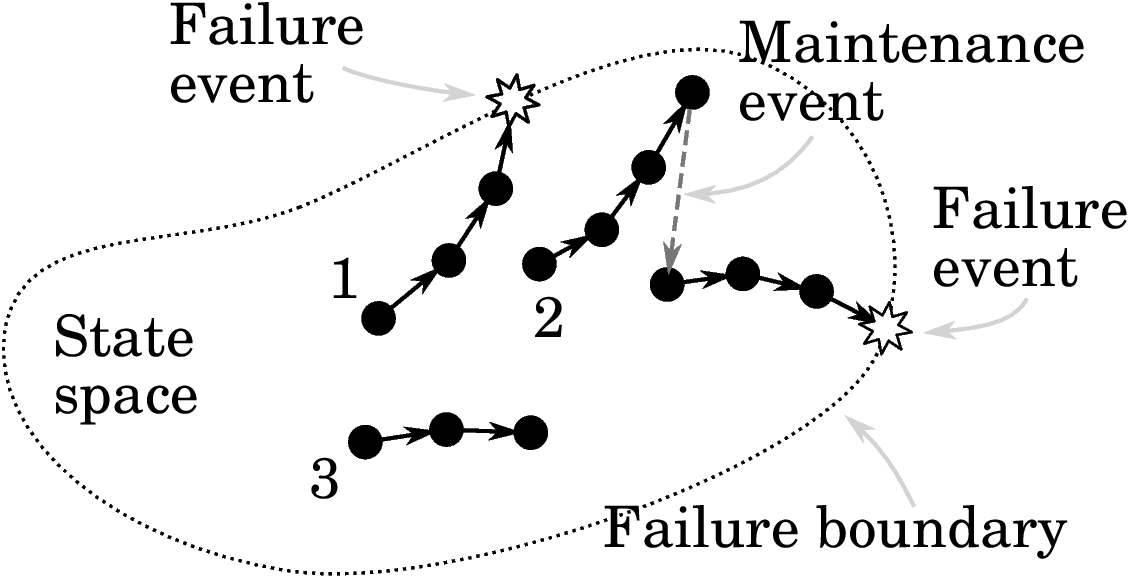}
\caption{Illustration of state-space trajectories for 3 assets. Solid circles indicate measurement events, stars represent failure events. The dash arrow indicates a state transition due to maintenance action.}\label{fig:statespace}
\end{figure}

Figure~\ref{fig:statespace} illustrates this concept. The trajectories of three assets in the latent degradation space are shown. The failure boundary reflects the level of degradation that results in an observed failure. Asset 1 degrades and fails. For this asset, time-till-failure would be retrosepctively available. Asset 2 degrades, but transitions to another point of the state-space due to maintenance action, after which it degrades to failure along a different trajectory. For this asset time-till-failure would be misleading for early observations as they are confounded by the effect of the maintenance. No failure is observed for Asset 3, making it unusable for simple supervised \ac{RUL}-based methods. 

Figure~\ref{fig:statespace} represents a Markovian view-point.
But the existence of such a latent state is well motivated by the predictive state representation (\ac{PSR}) approach to partially observable Markov decision processes (\ac{POMDP}s)~\cite{littman2002predictive,singh2003learning,boots2011closing}. Our perspective is that modeling the degradation process in this way neatly addresses the effects of maintenance on system evolution towards failure.
We can structure this formulation as a representation learning task. The objective would be to learn an embedding function that would map an asset's history into a latent vector representation. The evolution of these vectors could be presumed (e.g.\ incremented by cumulative historical load), or modeled. Finally, the effect of each maintenance action could also be modeled. 
This approach naturally makes use of all available data whether or not failures are observed. It can be realized in many ways. For example, one could use deep recurrent networks to map histories to the latent state and model maintenance as additive functions of current state and action type.
Finally, structural, physical, or reliability models of the assets can be incorporated into this modeling exercise to reduce data-driven model learning costs and improve accuracy.

\section{Conclusion}\label{sec:conclusion}
We reviewed current literature in the field of predictive maintenance. We identified several fundamental differences between condition estimation and failure risk forecasting as applied to simple components such as bearings and cutting tools from the capabilities needed to solve the same tasks on complex assets. These differences stem from complex latent degradation states, active maintenance programs, increased coupling between maintenance actions, and higher monetary and safety costs for failures.

As a result, methods that are effective for forecasting risk and informing maintenance decisions for individual components do not readily scale to sub-system or system level insights. A holistic modeling approach is needed that incorporates available structural and physical knowledge and naturally handles the complexities of actively fielded and maintained assets.

\bibliographystyle{apalike}

\appendix
\section{Acronyms}
\printacronyms[heading=none,sort=true]

\end{document}